%% file: main.tex
\documentclass[runningheads]{llncs}
\usepackage{graphicx}
\usepackage{hyperref}
\usepackage{amsmath}
\usepackage{amsfonts}
\usepackage{ctable}
\usepackage{bbm}
\usepackage{multirow}
\usepackage{xcolor,colortbl}
\usepackage[labelfont=bf]{caption}
\usepackage{subcaption}

\hypersetup{
colorlinks=true,
citecolor=blue,
urlcolor=red,
}

\begin{document}
\title{OOOE: Only-One-Object-Exists Assumption to Find Very Small Objects in Chest Radiographs} 
\titlerunning{OOOE: Only-One-Object-Exists Assumption}
%
\author{
    Gunhee Nam \and
    Taesoo Kim \and
    Sanghyup Lee \and
    Thijs Kooi
}
\authorrunning{Nam et al.}
%
\institute{Lunit Inc.\\
\email{\{ghnam, taesoo.kim, eesanghyup, tkooi\}@lunit.io}}
%
\maketitle              
\input{utils/macros}
\input{utils/symbols}

\input{utils/abbr}

\input{body/00_abstract}
\input{body/01_intro}

\input{body/03_methods_thijs}
\input{body/04_experiment}

\input{body/05_conclusion}

\input{body/06_acknowledgement}
\bibliographystyle{splncs04}
\bibliography{refs}

\end{document}


\title{Supplementary Materials \\
\small{OOOE: Only-One-Object-Exists Assumption \\ to Find Very Small Objects in Chest Radiographs}
        } 
\titlerunning{OOOE: Only-One-Object-Exists Assumption}
%
\author{}
\authorrunning{XXX et al.}
%
\institute{}
\maketitle              

\input{supple/figures/carina_anno}
\input{supple/figures/precision_plots_abs_dist}
\input{supple/figures/precision_plots}
\input{supple/tables/imp_details}

%

%% file: utils/macros.tex
\newcommand{\etal}{\textit{et al}.}

\newcommand{\eg}{\textit{e.g.~}}
\newcommand{\ie}{\textit{i.e.~}}

\newcommand{\tref}[1]{Tab.~\ref{#1}}
\newcommand{\Tref}[1]{Table~\ref{#1}}
\newcommand{\eref}[1]{Eq.~(\ref{#1})}
\newcommand{\Eref}[1]{Equation~(\ref{#1})}
\newcommand{\fref}[1]{Fig.~\ref{#1}}
\newcommand{\Fref}[1]{Figure~\ref{#1}}
\newcommand{\sref}[1]{Sec.~\ref{#1}}
\newcommand{\Sref}[1]{Section~\ref{#1}}

\newcommand{\dummyfig}[1]{
  \centering
  \fbox{
    \begin{minipage}[c][0.33\textheight][c]{0.5\textwidth}
      \centering{#1}
    \end{minipage}
  }
}

%% file: utils/symbols.tex
\newcommand{\img}{\mathbf{I}}
\newcommand{\pred}{\mathbf{y}}
\newcommand{\featureMap}{\mathbf{X}}
\newcommand{\feature}{\mathbf{x}}
\newcommand{\featureExtractor}{F}
\newcommand{\head}{g}
\newcommand{\sigmoid}{\sigma_p}
\newcommand{\spatialSoftmax}{\sigma_s}
\newcommand{\largeset}{S$_\text{train}$}
\newcommand{\smallset}{S$_\text{test}$}

\newcommand{\Exp}{$E$}
\newcommand{\IIExp}{\Exp$_{I \rightarrow I}$}
\newcommand{\RRExp}{\Exp$_{R \rightarrow R}$}
\newcommand{\IRExp}{\Exp$_{I \rightarrow R}$}
\newcommand{\IRExpPlus}{\Exp$_{I \rightarrow R^+}$}
\newcommand{\RIExp}{\Exp$_{R \rightarrow I}$}

\newcommand{\IIExpTT}{\Exp$_{\text{\notube} \rightarrow \text{\notube}}$}
\newcommand{\IRExpTT}{\Exp$_{\text{\notube} \rightarrow R}$}
\newcommand{\notube}{$I^-$}

\newcommand{\argmin}{\mathop{\mathrm{argmin}}\limits}
\newcommand{\argmax}{\mathop{\mathrm{argmax}}\limits}

%% file: utils/abbr.tex
\newcommand{\ett}{ETT}

%% file: body/00_abstract.tex
\begin{abstract}
The accurate localization of inserted medical tubes and parts of human anatomy is a common problem when analyzing chest radiographs and something deep neural networks could potentially automate. 
However, many foreign objects like tubes and various anatomical structures are small in comparison to the entire chest X-ray, which leads to severely unbalanced data and makes training deep neural networks difficult. 
In this paper, we present a simple yet effective `Only-One-Object-Exists' (OOOE) assumption to improve the deep network's ability to localize small landmarks in chest radiographs. 
The OOOE enables us to recast the localization problem as a classification problem and we can replace commonly used continuous regression techniques with a multi-class discrete objective. We validate our approach using a large scale proprietary dataset of over 100K radiographs as well as publicly available RANZCR-CLiP Kaggle Challenge dataset and show that our method consistently outperforms commonly used regression-based detection models as well as commonly used pixel-wise classification methods. Additionally, we find that the method using the OOOE assumption generalizes to multiple detection problems in chest X-rays and the resulting model shows state-of-the-art performance on detecting various tube tips inserted to the patient as well as patient anatomy.

\keywords{Point detection  \and Localization \and Object segmentation.}
\end{abstract}

%% file: body/01_intro.tex
\section{Introduction}

A common and effective application of deep neural networks in the domain of automated Chest X-ray (CXR) analysis is the localization of foreign objects and human anatomy \cite{CALLI2021}.
For example, the ability to segment and locate foreign objects, such as catheters, tubes, and lines has tremendous potential to optimize clinical workflow and ultimately improve patient care \cite{godoy2012chest1,godoy2012chest2,gupta2014postprocedural}. The innovations in object detection and segmentation methods for natural images~\cite{girshick2015fast_rcnn,lin2017fpn,ran2015faster_rcnn,ronneberger2015unet} have sparked progress in detecting foreign objects and anatomy in CXR images \cite{frid2019simul,Lee2017ADS,sullivan2020deep}. However, despite unique challenges associated with finding objects in CXR images, many of the methods designed for equivalent tasks in natural images are applied to CXR images without significant architectural modifications.

Compared to most objects in natural images, foreign objects and human anatomy viewed in CXR images are much smaller in scale. 
Training deep neural network to detect small scale objects is challenging, because the number of background pixels far outweighs the foreground pixel count \cite{chen2016r,lin2017focal,tong2020recent}. 
Frid-Adar \etal~\cite{frid2019simul} proposed to generate training data by synthesizing images with augmented endotracheal tubes (ETT). Their method addresses data imbalance and improves performance of an image level classification, but does not provide a solution for the small object detection problem. 
Kara \etal \cite{kara2021ett} proposed a regression based cascade method to localize the tip of ETT and the carina. In comparison, we provide a classification based solution to the detection problem which is often reported to outperform regression based methods for various detection tasks in natural images~\cite{jakab2018unsup_keypoint,li2018bin_cls,li2017kp_cls,oh2017color,su20153d_cls}. 

In this work, we present a solution to the problem of detecting small foreign objects or anatomical structures in chest radiographs. We introduce the \textit{`Only-One-Object-Exists'} (OOOE) assumption, a simple yet effective assumption, that limits the number of observable instances of a particular object we want to detect to one per image and reduces the detection problem to a point localization problem. Using these assumptions, the localization problem can be cast as a classification problem that can be solved with a spatial-softmax operation. 


We validate our approach for (1) detecting ETT tip and (2) detecting the carina, on the publicly available RANZCR-CliP Kaggle Challenge dataset. Additionally, we also provide results on a large scale proprietary dataset of over 100K chest X-ray images. Our method inspired by the OOOE assumption outperforms two commonly used baselines: (1) a simple segmentation model \cite{long2015fcn,ronneberger2015unet} and (2) a regression based detection approach \cite{kara2021ett}.  We additionally demonstrate that our approach leads to a model that generalizes better across datasets and makes better use of global context information.
  


%% file: body/03_methods_thijs.tex
\section{Methods}

\input{resources/figures/head_outputs}
We address the problem of detecting small objects in an image, using the assumption that they occur once and only once. We also observe that small objects, such as the tip of a tube or a certain landmark of an anatomy, can essentially be represented as a single point in an image.

Our solution to the point detection problem consists of two parts: a feature extractor $\featureExtractor$ and a detection head $\head$, which will be described in detail in the following sections. 

\subsection{Feature Extractor}
A feature extractor is a function that satisfies the following: 
\begin{equation}
    X = \featureExtractor(\img),
\end{equation}
where $\img \in \mathbb{R}^{H \times W \times C}$ is an input image with spatial dimensions $H,W$ and with $C$ channels. The feature extractor $\featureExtractor$ is a transformation such that the output feature $X \in \mathbb{R}^{h \times w \times c}$ is a tensor with spatial dimensions $h,w$ such that $h<H, w<W$ and with $c$ channels. In this work, we implement $\featureExtractor$ with a widely used convolutional neural network with residual connections (\texttt{ResNet34}) \cite{he2016resnet}. 

\subsection{Point Detection Head}
\label{sec:regression}
In a point detection problem, we assume that the ground truth location of an object of interest is represented as a single 2D location on a image $\pred \in \mathbb{R}^2$.
The objective of the point detection head $\head$ is to predict $\hat{\pred}$ given $X$. Depending on how  $\hat{\pred}$ is computed and how $\head$ is trained, a detection algorithm is considered to be either a regression or a classification method. 

Regression based approaches such as \cite{kara2021ett} are trained by directly minimizing the mean-square-error (MSE) between the predicted location $\head(X) = \hat{\pred}_{reg} \in \mathbb{R}^2$ and the ground truth location $\pred$ as depicted in (a) of Figure \ref{fig:head_outputs}. 

Despite the simplicity of regression based detection methods, classification based methods have outperformed them in practice across multiple detection problems ~\cite{jakab2018unsup_keypoint,jeon2020pose_att,li2018bin_cls,li2017kp_cls,oh2017color,su20153d_cls}. In a classification setup, the model instead outputs an activation map $\head(X) = \hat{\pred} \in \mathbb{R}^{H \times W}$ where the value located at $(i, j)$ is $\hat{\pred}_{ij} = \head(\feature_{ij})$ and $\featureExtractor(\img) = \featureMap = [\feature_{11}, \feature_{12}, ... , \feature_\textit{WH}]$. As shown in (b) of Figure \ref{fig:head_outputs}, the presence of an object at $\hat{\pred}_{ij}$ is learned by computing a \textit{pixel-wise binary cross-entropy} (BCE) loss with $\pred_{ij}$ where $\pred_{ij}=1$ when the ground truth location of the object is at $i,j$ and $\pred_{ij}=0$ otherwise. 

One of the main intuition of this paper is that we can often and naturally bound the number of positive detections in $\hat{\pred}$ by using application driven prior knowledge. It is often true for practical applications that the expected number of object/anatomy is known a priori and is equal to one (eg. humans only have one carina, only one endo-tracheal tube is inserted at any given time). Our idea is to encode this strong prior using the \textit{spatial softmax} operator (as opposed to pixel-wise BCE) which leads to our OOOE point detection head formulation which we describe below.


\subsubsection{Spatial softmax}

The spatial softmax allows the detection head to produce a relative probability for each pixel, by applying the softmax function along the spatial axis. This leads to the OOOE assumption we make in this paper which states one and only one instance of the object is present in the image. 

The spatial softmax $\spatialSoftmax$ operation over the activation map $\hat{\pred}$ and the resulting value at spatial location $i,j$ is defined as follows:
\begin{equation}
    \spatialSoftmax(\hat{\pred}_{ij}) = \frac{e^{\hat{\pred}_{ij}}}{\sum_{j=1}^H\sum_{i=1}^{W} e^{\hat{\pred_{ij}}}}.
\end{equation}
Then, the point detection model is optimized to minimize the following negative log-likelihood objective:
\begin{equation}
    \mathcal{L}(\pred, \hat{\pred}) = \frac{1}{HW}\sum_{j=1}^{H}\sum_{i=1}^{W} - \pred_{i,j} \log (\spatialSoftmax(\hat{\pred}_{ij})),
\end{equation}
where $\pred_{i,j} = 1$ when the object is located at point $(i,j)$. The final point detection prediction $\hat{\pred}_{cls}$ using classification based the spatial softmax approach is defined as the location $(i,j)$ with the highest activation value $\hat{\pred}_{ij}$:
\begin{equation}
    \hat{\pred}_{cls} = \argmax_{i,j} \hat{\pred}_{ij}
\end{equation}

Visual comparison to the regression and pixel-wise classification approaches is depicted in (c) of Figure \ref{fig:head_outputs}. 

%% file: resources/figures/head_outputs.tex
\begin{figure*}[t]
\centering
    \includegraphics[width=1.0\textwidth]{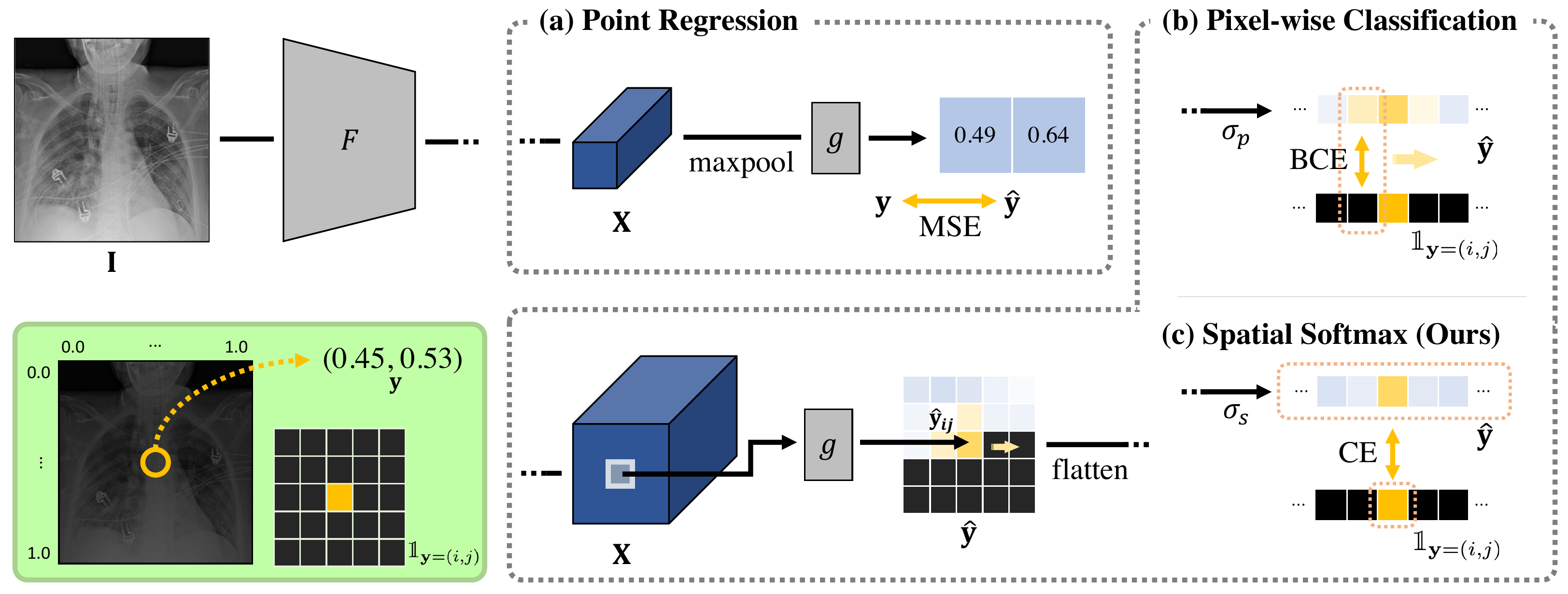}
    \caption{Comparison to regression and pixel-wise classification methods. Given a CXR image $\img$, our objective is to find the ground-truth $\pred$ point location.
            (a) The regression method predicts $\hat{\pred} = [\hat{\pred}_1,\hat{\pred}_2]$ directly.
            (b) The pixel-wise classification method predicts absence/presence of an object for all pixels.
            (c) Our method assumes only one instance of an object exists in an image. }
    \label{fig:head_outputs}
\vspace{-15pt}
\end{figure*}

%% file: body/04_experiment.tex
\section{Experiments}

\subsection{Datasets}
\input{resources/tables/data}
We evaluate the point detection performance on a relatively small public RANZCR-CLiP~\cite{ranzcr-clip} dataset and a large internal dataset. 
The differences are noted in \Tref{table:data}.
For both datasets, we define two subsets~(\largeset, \smallset) and cases in each split are randomly selected without patient id overlap between the splits.

\textbf{RANZCR-CLiP}~\cite{ranzcr-clip} is a dataset used in a recent Kaggle challenge for malpositioning classification of endotracheal and nasogastric tubes, and catheters. This dataset consists of 30K cases with case-level labels. Additionally, tube line annotations for a subset of (\~{}3K cases for ETT) are provided.
Using these line annotations, we create a ETT tip point annotation by taking the bottom most point in the ETT line annotation as the tip point.
For the same dataset, we use the trachea bifurcation~(\ie carina) point annotations provided by Konya \etal~\cite{carina}.

\textbf{Internal Dataset} refers to a large proprietary dataset.
The cases are collected from public data ~\cite{bustos2020padchest,irvin2019chexpert,johnson2019mimic,wang2017cxr14} as well as private sources consisting of various sites in multiple countries. 
Most of the cases are antero-posterior~(AP) images since the cases with tube objects are mostly from ill patients in a bedridden state.
For the dataset, 100K cases are annotated by 20 board-certified radiologists with previous CXR annotation experience.\footnote{Unfortunately, we are not in the position to disclose this data at this time.}

\subsection{Evaluation Metrics}
Previous studies for carina and ETT tip detection adopt an \textit{absolute error} to measure the model performance~\cite{kara2021ett}. Various statistics  such as the mean, median and standard deviation are reported. However, some of these statistics are sensitive to outliers.

In this paper, we additionally use precision plots which is a general metric to evaluate point detection performance. These plots are commonly used in object tracking literature~\cite{babenko2010tracking2,wu2013otb,henriques2012tracking1}. The precision plot shows the percentage of cases where the location error between the prediction and ground-truth is within a distance threshold $\delta$ on the y-axis against multiple prediction thresholds on the x-axis. This method reduces the effect of outliers, so that overall performance can be seen without severe bias.

Since the RANZCR-CLiP data does not provide information about the pixel spacing of the radiograph, we measure the distance relative to the size of the image and choose the maximum distance threshold $\delta = 0.15$. To summarize the performance, we report area-under-curve~(AUC) of the precision plots. We made use of bootstrapping to generate confidence bounds around the AUC values.

In \smallset~of the internal data, however, the DICOMs of some cases~(1,413 cases for carina, 524 cases for ETT tip) have pixel spacing information so that the absolute distance can be retrieved. 
To compare our method to related work, we report the same statistics used in \cite{kara2021ett} ~(\eg mean, median, etc.) of absolute errors for these cases including AUC of the precision plots.
For the precision plots by the absolute distance, we choose the maximum distance threshold $\delta = 50mm$.

\subsection{Implementation Details}

For the \textit{pixel-wise classification} method, we balance weights between positive and negative samples with the same ratio.
Otherwise, the model too easily over-fits to negative samples given the severe data imbalance; only one pixel in an image is positive for the point detection.
For the \textit{regression} method, we choose a learning rate of 0.001 by grid hyper-parameter search.

\input{resources/tables/exp_settings}
\input{resources/tables/performance_t}
\input{resources/tables/abs_dist}

\textbf{Experimental Settings}:
We trained and validated the model on different permutations of the data, described in \Tref{table:exp_setting}. 
For \IIExp~and \RRExp~ settings, we train and validate on the cases from the same source. Furthermore, we defined \IRExp ~and \RIExp~ experiments to test our model's ability to generalize across different data sources. Given this setup, we can also observe the effect of training set size on model's performance. 

\subsection{Comparison to other methods}
First, we compare our model to the regression based approach \cite{kara2021ett} in \Tref{table:performance}.
Our spatial softmax method outperforms all other methods across all settings. 

\input{resources/figures/carina_anntation_main}
The performance of carina detection is actually relatively worse when trained on a larger dataset (ie. \RRExp ~$>$ \IRExp). We suspect this is an effect of domain gap; there exists annotation style difference between the two datasets as shown in \Fref{fig:carina_anno}. 

\Tref{table:abs_dist} summarizes various statistical measures including the AUC of the precision plots of the absolute distance errors on \IIExp. 
Overall, our method shows the best performance compared to the other methods when measuring performance with respect to absolute distance error.

\subsection{A Closer Look at ET-tube vs. T-tube Detection Performance}
Upon qualitative analysis of our model's performance on the RANZCR-CLiP dataset, some cases classified as ETT actually turned out to be tracheostomy tubes (TT). TT is a short, curved airway tube that is inserted through a surgically generated stoma at the anterior neck, for prolonged respiratory support. TT is visually similar to ETT but TT can be discriminated from ETT by its typical course and short length. 



When looked at with a limited field of view, the TT and ETT are very similar as shown in \Fref{fig:ttube}. To discriminate ETT from TT, the model should take the context into account and look at the whole scan. Since the spatial softmax method compares relative scores from all pixels in the image, we postulate it is better at discriminating between ETT and TT than other methods. This is a highly desirable trait for a tube detection model as reducing such false positive cases not only improves performance but also even contribute to getting the trust of users of automated detector in practice. 

To test this hypothesis, we excluded cases with TT annotations from our internal dataset (\notube). The resulting performance of the different models is shown in
\Tref{table:ttube}. Our spatial softmax method indeed outperforms the pixel-wise classification method for \IIExpTT. On the other hand, our method shows much lower performance for its ETT tip detection performance than \IRExp~in \Tref{table:performance} while the pixel-wise classification method achieves relatively consistent performance. This indicates that the spatial softmax method is able to discriminate ETT from TT while the pixel-wise classification method does not.

In addition to precision, we also report the AUC of the receiver operating characteristic curve~(AUCROC) on \IIExpTT~in \Tref{table:ttube} for detecting ETT tips. Here, positives are cases with an ETT and negatives are cases with an TT tube, respectively. We used the maximum score from the prediction map as the prediction score.
The spatial softmax method outperforms the pixel-wise classification method by a large performance gap, which was found to be significant using a DeLong test ($p=0.0007$). 

\input{resources/tables/ttube}

%% file: resources/tables/data.tex
\begin{table*}[t]
\setlength{\tabcolsep}{15pt}
\centering
\caption{Dataset configuration. Each dataset is randomly split into two subsets~(\largeset, \smallset) with different sizes.}
\label{table:data}
\begin{tabular}{l|c|cc||c}
\toprule
\multicolumn{2}{l|}{}                    &  Carina &  ETT tip &   Total \\
\midrule
\multirow{2}{*}{RANZCR-CLiP~\cite{ranzcr-clip}}   &      \largeset &    4,244 &         2,057 &    5,931 \\
                                                  &       \smallset &    1,031 &          937 &    1,818 \\

\midrule

\multirow{2}{*}{Internal Dataset}      &      \largeset &   98,382 &        43,066 &  103,394 \\
                               &      \smallset &    1,544 &          609 &    1,633 \\
\bottomrule
\end{tabular}
 \vspace{-15pt}
\end{table*}

%% file: resources/tables/exp_settings.tex
\newcommand{\tr}{$\diamondsuit$}
\newcommand{\te}{$\star$}

\begin{table*}[t]
\centering
\footnotesize
\caption{Experiment configurations. Each experiment~(\Exp) uses different set for its training and test set. Diamond~(\tr) represents training set, and star(\te) represents test set for the experiment.}
\setlength{\tabcolsep}{23pt}
\label{table:exp_setting}
\begin{tabular}{l|c|c||c|c}
\toprule
\multirow{2}{*}{Split} & \multicolumn{2}{c||}{RANZCR-CLiP~\cite{ranzcr-clip}} & \multicolumn{2}{c}{Internal Dataset}  \\
\cmidrule{2-5}
 &       \largeset &       \smallset &    \largeset &       \smallset \\
\midrule
\IIExp &           &          &       \tr &         \te \\
\RRExp    &         \tr &        \te &       &        \\
\IRExp    &           &        \te &       \tr &         \\
\RIExp     &         \tr &         &       &    \te     \\
\bottomrule
\end{tabular}
 \vspace{-5pt}
\begin{flushright}\scriptsize{\tr~: Training set, \te~: Test set}\end{flushright}
 \vspace{-25pt}
\end{table*}

%% file: resources/tables/performance_t.tex
\begin{table}[t]
\setlength{\tabcolsep}{2pt}
\centering
\caption{Model performance for various experimental settings (see \Tref{table:exp_setting}). We report the area under the curve of precision plots~(Prec.) and its 95\% confidence interval~(95\% CI). Our spatial softmax~(SS) method outperforms other methods in all experimental settings.}
\label{table:performance}
\begin{tabular}{l|l||c|c||c|c||c|c||c|c}
\toprule
\multicolumn{2}{c||}{} & \multicolumn{2}{c||}{\IIExp} & \multicolumn{2}{c||}{\RIExp} & \multicolumn{2}{c||}{\RRExp} & \multicolumn{2}{c}{\IRExp} \\
\cmidrule{3-10}
\multicolumn{2}{r||}{(\%)} &          Prec. &       95\% CI &            Prec. &     95\% CI &       Prec. &     95\% CI &       Prec. &     95\% CI \\
\midrule
\multirow{3}{*}{Carina}  &        Reg &         74.5 &  73.7-75.3 &           46.5 &  45.2-47.7 &      59.2 &  57.8-60.6 &      69.1 &  68.2-70.2 \\
&         PC &         85.7 &    85.0-86.4 &           71.4 &  70.6-72.3 &        87.0 &  86.4-87.5 &      79.2 &  78.4-79.8 \\
\cmidrule{2-10}
&        \textbf{Ours} &         \textbf{86.8} &    86.1-87.5 &           \textbf{72.1} &  71.1-72.9 &      \textbf{89.3} &  88.7-89.9 &      \textbf{81.1} &  80.7-81.6 \\
\midrule
\multirow{3}{*}{ETT Tip} &        Reg &         69.1 &    67.1-70.9 &           35.1 &  32.8-37.9 &      42.4 &  41.0-44.0 &      67.2 &  65.5-68.8 \\
&         PC &         82.9 &    81.1-84.3 &           60.8 &  58.7-63.6 &      67.5 &  65.8-69.1 &      75.1 &  73.1-76.9 \\
\cmidrule{2-10}
&         \textbf{Ours} &         \textbf{87.4} &    86.3-88.8 &           \textbf{73.6} &  71.7-75.3 &      \textbf{70.4} &  68.8-71.8 &      \textbf{76.8} &  74.9-78.7 \\
\bottomrule
\end{tabular}
\vspace{-5pt}
\begin{flushright}\scriptsize{\textbf{Reg}: Regression, \textbf{PC}: Pixel-wise Classification, \textbf{Ours}: Spatial Softmax}\end{flushright}
\vspace{-25pt}
\end{table}

%% file: resources/tables/abs_dist.tex
\begin{table}[t]
\centering
\footnotesize
\caption{Absolute Distance Error between Predictions and Ground-truths. }
\setlength{\tabcolsep}{2.5pt}
\label{table:abs_dist}
\begin{tabular}{l|l||ccccccc|c||c}
\toprule
\multicolumn{2}{l||}{Error (mm)} &   Mean &  Median &     Max &   Min &    Std &    Q1 &     Q3 &  Prec.(\%) &   count \\
\midrule
\multirow{3}{*}{Carina} & Reg             &  14.46 &   12.51 &  107.00 &  0.04 &   9.75 &  7.70 &  18.34 &          71.29 &  \multirow{3}{*}{1,413} \\
 & PC    &   7.81 &    5.64 &   \textbf{86.59} &  0.01 &   \textbf{7.64} &  3.19 &   9.64 &          84.40 & \\
\cmidrule{2-10}
 & \textbf{Ours}         &   \textbf{7.46} &    \textbf{4.75} &  229.51 &  \textbf{0.00} &  10.58 &  \textbf{2.66} &   \textbf{8.71} &          \textbf{85.56} & \\
\midrule
\multirow{3}{*}{ETT Tip} & Reg           &  18.38 &   13.72 &  141.38 &  \textbf{0.00} &  15.97 &  8.34 &  23.54 &          65.00 &   \multirow{3}{*}{524} \\
 & PC &   9.73 &    5.46 &  128.32 &  \textbf{0.00} &  13.33 &  3.00 &   9.53 &          81.41 &  \\
\cmidrule{2-10}
 & \textbf{Ours}      &   \textbf{7.28} &    \textbf{3.95} &   \textbf{98.75} &  \textbf{0.00} &  \textbf{11.80} &  \textbf{2.40} &   \textbf{6.64} &          \textbf{86.28} &  \\

\bottomrule
\end{tabular}
 \vspace{-5pt}
\end{table}

%% file: resources/figures/carina_anntation_main.tex
\begin{figure*}[h]
    \centering
    \includegraphics[width=1.0\textwidth]{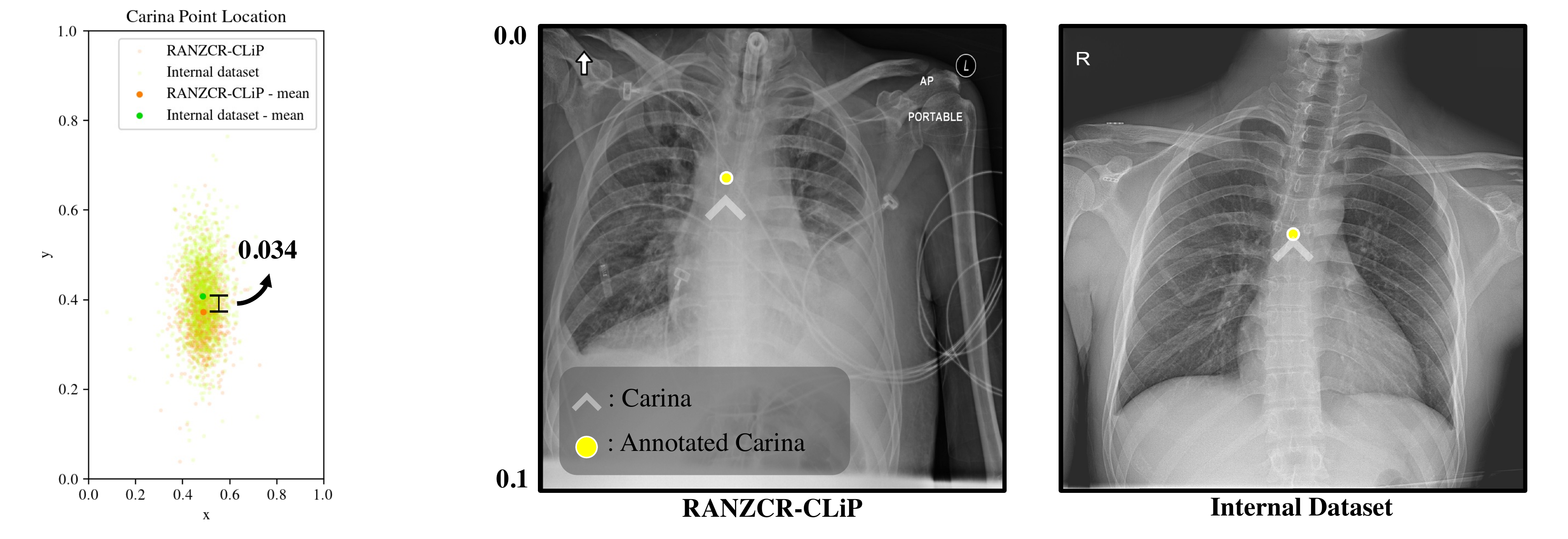}
    \vspace{-20pt}
    \caption{Between the annotations from RANZCR-CLiP and the internal dataset, the intrinsic annotation inconsistency exists. \textit{(Left)} The carina point location plot of \smallset~in each dataset and its mean value. The mean location difference in relative distance is 0.034 between the two dataset. \textit{(Right)} The visualization of the annotated carina and the actual carina point for each dataset. The carina is annotated slightly upper than the actual point in RANZCR-CLiP.}
    \label{fig:carina_anno}
    \vspace{-15pt}
\end{figure*}

%% file: resources/tables/ttube.tex
 \begin{table}[t]
  \centering
  \begin{minipage}[t]{.6\linewidth}
    \vspace{5pt}
    
    \centering
    \includegraphics[width=1.0\textwidth]{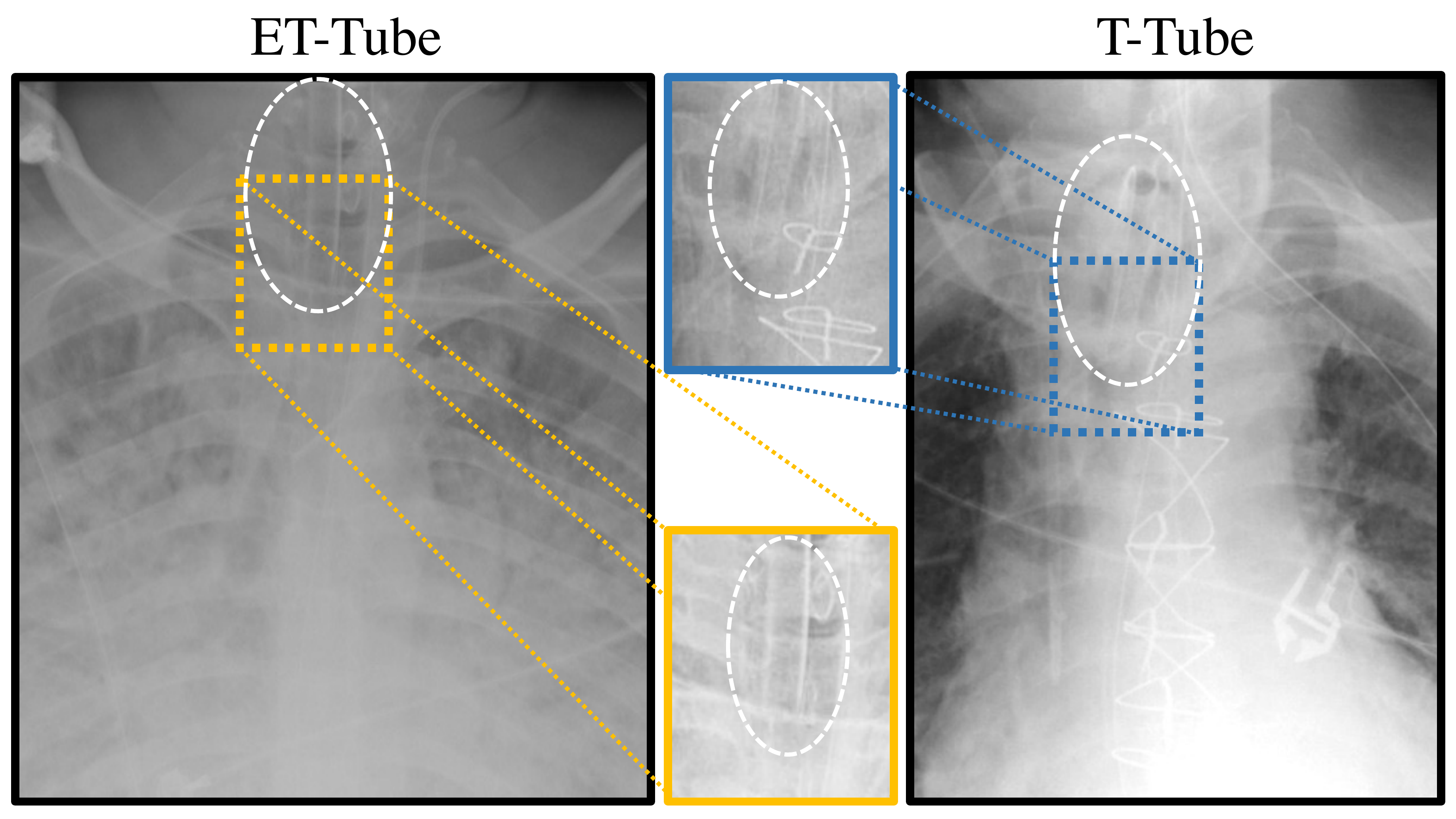}
    \vspace{-10pt}
    \captionof{figure}{ETT and TT in CXR. The local appearances are so similar that it is hard to discriminate between the two without the global view.}
    \label{fig:ttube}
  \end{minipage}
  \hfill
  \begin{minipage}[t]{.35\linewidth}
    \vspace{0pt}
    
    \setlength{\tabcolsep}{3pt}
    \captionof{table}{AUC of precision plots~(Prec.) and AUCROC~(AUC) for ETT tip vs. TT. \notube~denotes the internal dataset excluding TT from ETT annoations.}
    \label{table:ttube}
    \begin{tabular}{l|c||c|c}
    \toprule
    \multirow{2}{*}{(\%)} &   \IRExpTT &   \multicolumn{2}{c}{\IIExpTT} \\
    \cmidrule{2-4}
    &  Prec. &  Prec. &     AUC \\
    \midrule
    Reg &       55.3 &       61.4 &          - \\
    PC  &       \textbf{67.8} &       83.5 &       80.6 \\
    \cmidrule{1-4}
    \textbf{Ours}  &       63.2 &       \textbf{87.1} &       \textbf{88.3} \\
    \bottomrule
    \end{tabular}
  \end{minipage}\hfill
  \vspace{-25pt}
\end{table}

%% file: body/05_conclusion.tex
\section{Conclusion}
In this paper, we presented a method for the detection of small single objects in medical images, inspired by work from landmark detection in natural images. The method is simple to implement and outperforms other commonly used techniques such as methods based on regression or pixel-wise segmentation by a large margin on two different detection tasks in chest X-ray, using two different datasets. Although we prove the effectiveness of our method in CXR images, the detection of small single objects for other modality remains undiscovered. 
This may have potential to expand into other medical imaging areas, such as detecting a clip markers in mammograms. As part of future work, the method could be extended to assume a variable number of instances (\eg 2 or more) and would increase the range of possible applications. We hope an expert knowledge driven automated system as presented in this paper contributes to increased application of automated methods in real world practice. 

\textbf{Prospect of application:}
The ability to detect the position of inserted endo-tracheal tube tip with respect to the patient's carina from chest X-rays has the potential to enable malpositioning detection of the tube. 

%% file: body/06_acknowledgement.tex
\footnotesize
\textbf{Acknowledgement} This version of the contribution has been accepted for publication, after peer review but is not the Version of Record and does not reflect post-acceptance improvements, or any corrections. The Version of Record is available online at: \url{https://doi.org/10.1007/978-3-031-17721-7_15}. Use of this Accepted Version is subject to the publisher's Accepted Manuscript terms of use \url{https://www.springernature.com/gp/open-research/policies/accepted-manuscript-terms}.

%% file: supple_/figures/carina_anno.tex
\begin{figure*}[h]
    \vspace{-30pt}
    \centering
    \includegraphics[width=1.0\textwidth]{supple/figures/resources/carina_anno.pdf}

    \caption{Between the annotations from RANZCR-CLiP and the internal dataset, the intrinsic annotation inconsistency exists. \textit{(Left)} The carina point location plot of \smallset~in each dataset and its mean value. The mean location difference in relative distance is 0.034 between the two dataset. \textit{(Right)} The visualization of the annotated carina and the actual carina point for each dataset. The carina is annotated slightly upper than the actual point in RANZCR-CLiP. This is consistent to the observation from the left plot. We suspect this difference causes the performance drop even when trained with the larger dataset in Table 3 in the main paper.}
    \label{fig:carina_anno}
    \vspace{-25pt}
\end{figure*}

%% file: supple_/figures/precision_plots_abs_dist.tex
\begin{figure*}[h]
    \centering
    \begin{subfigure}[b]{0.49\textwidth}
        \centering
        \includegraphics[width=1.0\textwidth]{supple/figures/resources/precision_plots_for_Carina.png}
    \end{subfigure}
    \begin{subfigure}[b]{0.49\textwidth}
        \centering
        \includegraphics[width=1.0\textwidth]{supple/figures/resources/precision_plots_for_ETTubeTip.png}
    \end{subfigure}
    \caption{Precision plots for carina and ETT tip according to the absolute distance error in Table 4 of the main paper.}
    \label{fig:precision_plots_abs_dist}
\end{figure*}

%% file: supple_/figures/precision_plots.tex
\begin{figure*}[t]
    \centering
    \begin{subfigure}[b]{0.49\textwidth}
        \centering
        \includegraphics[width=1.0\textwidth]{supple/figures/resources/internal_Carina.png}
    \end{subfigure}
    \begin{subfigure}[b]{0.49\textwidth}
        \centering
        \includegraphics[width=1.0\textwidth]{supple/figures/resources/internal_ETTubeTip.png}
    \end{subfigure}
    \begin{subfigure}[b]{0.49\textwidth}
        \centering
        \includegraphics[width=1.0\textwidth]{supple/figures/resources/large_Carina.png}
    \end{subfigure}
    \begin{subfigure}[b]{0.49\textwidth}
        \centering
        \includegraphics[width=1.0\textwidth]{supple/figures/resources/large_ETTubeTip.png}
    \end{subfigure}
    \begin{subfigure}[b]{0.49\textwidth}
        \centering
        \includegraphics[width=1.0\textwidth]{supple/figures/resources/small_Carina.png}
    \end{subfigure}
    \begin{subfigure}[b]{0.49\textwidth}
        \centering
        \includegraphics[width=1.0\textwidth]{supple/figures/resources/small_ETTubeTip.png}
    \end{subfigure}
    \begin{subfigure}[b]{0.49\textwidth}
        \centering
        \includegraphics[width=1.0\textwidth]{supple/figures/resources/small-econn_Carina.png}
    \end{subfigure}
    \begin{subfigure}[b]{0.49\textwidth}
        \centering
        \includegraphics[width=1.0\textwidth]{supple/figures/resources/small-econn_ETTubeTip.png}
    \end{subfigure}
    \caption{Precision plots for carina and ETT tip according to the relative distance error in Table 3 of the main paper. }
    \label{fig:precision_plots}
\end{figure*}

%% file: supple_/tables/imp_details.tex
\begin{table}[t]
\centering   

\setlength{\tabcolsep}{3pt}
\small
\caption{Implementation Details}
\label{table:imp_details}
\begin{tabular}{l|l||l|l}
\toprule
\textbf{Optimizer}     & SGD (momentum 0.9) &           \textbf{Backbone} &          \texttt{resnet34} + FPN  \\
\midrule
\textbf{Learning rate} &                                  0.04 &  \textbf{Init weight} &                ImageNet pretrained\\
\midrule
\textbf{Train epoch}   &                                    40 &               \textbf{GPUs} &  NVIDIA RTX 2080 Ti * 8 \\
\midrule
\textbf{LR drop} & $\times$ 0.1 at 30$^{\text{th}}$ epoch  &         \textbf{Batch size} &                     192 \\
\bottomrule
\end{tabular}
\end{table}